%% file: main.tex
\documentclass[sigconf, anonymous=false]{acmart}
\usepackage{algorithm}
\usepackage{subfigure}
\usepackage{algpseudocode}
\usepackage{amsmath}
\usepackage{graphicx}
\usepackage{float} 
\usepackage{colortbl}
\usepackage{arydshln}
\usepackage{multirow}
\usepackage{multicol}

\AtBeginDocument{%
  \providecommand\BibTeX{{%
    \normalfont B\kern-0.5em{\scshape i\kern-0.25em b}\kern-0.8em\TeX}}}

\setcopyright{acmcopyright}
\copyrightyear{2020}
\acmYear{2020}

\acmConference[KDD '27]{KDD '27: International Conference on Knowledge Discovery and Data Mining}{August 14--18, 2021}{Singapore, SG}
\acmBooktitle{KDD '27: International Conference on Knowledge Discovery and Data Mining,
  August 14--18, 2021, Singapore, SG}

\begin{document}

\title{Incorporating Vision Bias into Click Models for Image-oriented Search Engine}
\author{Ningxin Xu}
\email{ningxinx@andrew.cmu.edu}
\affiliation{%
  \institution{Carnegie Mellon University}
}

\author{Cheng Yang}
\email{yangcheng.iron@bytedance.com}
\affiliation{%
  \institution{ByteDance AI Lab, China}
}

\author{Yixin Zhu}
\email{zhuyixin@bytedance.com}
\affiliation{%
  \institution{ByteDance AI Lab, China}
}

\author{Xiaowei Hu}
\email{huxiaowei.01@bytedance.com}
\affiliation{%
  \institution{ByteDance AI Lab, China}
}

\author{Changhu Wang}
\email{wangchanghu@bytedance.com}
\affiliation{%
  \institution{ByteDance AI Lab, China}
}
\renewcommand{\shortauthors}{Ningxin, Cheng, and Yixin, et al.}

\begin{abstract}
Most typical click models assume that the probability of a document to be examined by users only depends on position, such as PBM and UBM. It works well in various kinds of search engines. However, in a search engine where massive candidate documents display images as responses to the query, the examination probability should not only depend on position. The visual appearance of an image-oriented document also plays an important role in its opportunity to be examined.
\par
In this paper, we assume that \textit{vision bias} 
exists in an image-oriented search engine as another crucial factor affecting the examination probability aside from position. Specifically, we apply this assumption to classical click models and propose an extended model, to better capture the examination probabilities of documents. We use regression-based EM algorithm to predict the \textit{vision bias} given the visual features extracted from candidate documents. 
Empirically, we evaluate our model on a dataset developed from a real-world online image-oriented search engine, and demonstrate that our proposed model can achieve significant improvements over its baseline model in data fitness and sparsity handling.
\end{abstract}



\maketitle
\section{Introduction}
\input{intro.tex}

\section{Related Works}
\input{rel.tex}

\section{Modeling}
\input{modeling.tex}

\section{Parameter Inference}
\input{infer.tex}

\section{Experiments}
\input{exp.tex}

\section{Conclusions}
In this paper, we propose \textit{vision bias} for the click model application working on image-oriented search engines, which captures the influence of the visual features on user examination probability. Few previous works have independently studied this topic and provided a specific solution in details. This paper makes up for this deficiency and provides a solution for documents retrieval in the image-oriented search engine by proposing a novel probabilistic graph structure for click models to incorporate the \textit{vision bias} into them. We assume that aside from position, visual features of image-oriented documents are another important factor that influences the examination probability of each document, and introduces a parameter representing \textit{vision bias} in the probabilistic graph. The assumptions about \textit{vision bias} are extensible, since it can be applied to the probabilistic graph of many classical click models such as PBM and UBM.
Parameters of existing click models can be estimated by standard EM algorithm. When we use a click model in the image-oriented search engine, visual features should be soundly utilized. Thus, we use regression-based EM algorithm and estimate \textit{vision bias} to obtain better performance, which also has an advantage in sparsity handling. 
\par
In the experiments, we evaluate our approach on the dataset obtained from an online image-oriented search engine. This dataset includes both click-through logs and candidate images. We extend PBM and UBM by modeling \textit{vision bias} as part of them. The experiment results demonstrate that the extended models are more effective for ranking and click prediction tasks. We also evaluate how well the extended models can handle the sparsity problem, and the experiment results show that in both data fitness and ranking effectiveness, extended models significantly outperform their baseline models on the sparse user data. 
\par
Since various deep learning techniques have been proposed to handle image features, future work can include more sophisticated neural network structure for \textit{vision bias} parameter inference in order to capture more demanded information from visual features.
\bibliographystyle{ACM-Reference-Format}
\bibliography{references}
\clearpage
\input{appendix.tex}
\end{document}

%% file: intro.tex
A commercial search engine provides services of retrieve and ranking for massive users. Improved ranking performance leads to desired results ranked higher, adds to user activity and makes profits for enterprises. Models have been designed to construct result pages to satisfy users' information needs more productively. In a commercial search engine, numerous click-through logs record user activities on search pages every day at low cost, but provide implicit, abundant and renewable user feedback. Many models have been proposed to take advantage of click-through logs to improve document rankings. These models are called \textit{click models}.In \cite{log}, click models are described as models that take noise and bias in user behavior data into account by introducing random variables and dependencies between them, and concern clicks as the main observed user interaction with a search system. According to \cite{log}, user browsing model (UBM) \cite{UBM}, dynamic Bayesian network model (DBN) \cite{DBN}, click chain model (CCM) \cite{CCM}, general click model (GCM) \cite{GCM}, position-based model (PBM) \cite{PBM} and cascade model (CM) \cite{PBM} are typical click models to utilize the abundant click-though logs.
\par
However, click models are faced with well-known challenges from the inherent biases contained in user behaviors. 
Examination-based click models such as PBM, CM and UBM have been proposed to estimate this bias to extract the real relevance of the documents. The more accurate relevance leads to better rankings. 
These click models have an excellent performance in search engines, but do not distinguish between different types of candidate documents such as image, text, links and etc. In recent years, search engines with images as candidate documents have emerged in large numbers, and the models now have to consider the nature of candidate documents. Compared with texts and links, images have strong visual impacts. In this case, the opportunity to be observed and examined of a candidate document more depends on the visual appearance of the image itself, not only on the position. Specifically, GIF images among static images, or images with abundant content aligned with monotonous ones, or images with bright colors compared with those in dim colors, are usually more likely to be observed and examined. Relevant proofs can be found in \cite{eyecatch}. 
In this case, the importance of focusing on this bias should be reflected in the design of user behavior modelings. In this paper, we define this kind of bias as \textit{vision bias}. 
\par
An intuitive approach to estimate the \textit{vision bias} caused by visual appearance is to utilize the information contained in the candidate image-oriented documents. 
This idea inspires us to propose an approach where a model captures the bias caused by visual features directly from the documents. 
In this paper, every image-oriented document is encoded with its feature vector, and infer its \textit{vision bias} by regression-based EM. Other parameters, such as relevance and position bias, are estimated via standard EM algorithm.
\par
 This paper incorporates the concepts of \textit{vision bias} to explain how user clicks are affected by visual features, and to extract more accurate document relevance to achieve better rankings. Click models which adopt the \textit{vision bias} can estimate more unbiased relevance in image-oriented search engines. Empirically, we apply the assumption of \textit{vision bias} to PBM and UBM, and evaluate the proposed method on click logs produced by real users from the online image-oriented search engine. Proposed model extracts the relevance between the query and the candidate documents, and then the results are compared with PBM and UBM on the test dataset.
\par
In summary, the main contributions of this paper are as follow:
\begin{itemize}
\item We propose \textit{vision bias} from click data that captures the influence of the visual features on the user examination when it comes to image-oriented search engines. 
\item We propose an reasonable equation to incorporate the \textit{vision bias} into click models by introducing a parameter representing it. In the experiments, we demonstrate that this equation is effective for better rankings.
\item We use regression-based EM algorithm to estimate \textit{vision bias}, and show that this can effectively handle data sparsity problem in click models.
\item Our method achieve high extensibility. \textit{Vision bias} can be added to the examination hypothesis of most of widely-used click models in image-oriented search engines. 
\end{itemize} 
\par
In the following sections, this paper discusses how to propose appropriate ways to combine position bias and vision bias, and demonstrates the advantage of this method. We review related works in Section 2, introduce the modeling approach in Section 3 and discuss the parameter inference method in Section 4. In Section 5 we present the results and analysis of our experiments. Finally, we conclude the paper in Section 6.

%% file: rel.tex
In the past, various biases have been addressed in typical click models, such as trust bias \cite{trust} and intent bias \cite{intent} \cite{intent2}. Other works, as presented in \cite{noise}, also incorporate inherent noise in the user behavior into click models. These works have studied factors that make click data noisy, and most of them focus on the influence of noisy click data on users' judgement on document relevance. This paper differs from them by introducing a bias into the examination instead of relevance. 
\par
Biases caused by what documents look like, known as appearance/presentation bias, have been studied in the previous works such as \cite{ClarkeSIGIR2007}, \cite{comment} and \cite{www2010bias}. The study \cite{ClarkeSIGIR2007} suggests the existence of bias and notices that when the title of documents have more matching terms with query, even though they are not very relevant, they get more clicks.  \cite{comment} incorporates the appearance/presentation bias into its model by introducing a parameter, and it assumes that this bias has impacts on the estimation of the relevance. In \cite{www2010bias}, it is pointed out that the "more attractive" results make the perceived relevance of some documents differ significantly from others. Our work differs from these papers by modeling a bias in the image-oriented search engines. We argue that \textit{vision bias} affects the user behavior by influencing the examination probability, and propose effective methods incorporating visual features to estimate it. 
\par
Work on using visual features for image search is long-standing, such as in \cite{vs1}, \cite{vs2}, \cite{vs3} and \cite{vs4}. \cite{dlimg} learns visual features in learning-to-rank tasks via deep learning techniques. \cite{imagesearchexamination} focuses on the examination behavior of image search users, and discovers that the content of image results (e.g.,visual saliency) affects examination behavior. It can be seen from these works that there are many techniques for image search which utilize visual features. In this paper we limit the research scope within click models, and extend classical click models by applying regression-based EM algorithm to them. Particularly, \cite{imgcm} proposes a novel interaction behavior model, grid-based user browsing model, which shares similarities with click model techniques, and this model works in image search. However, it does not explicitly model appearance/presentation bias, which is the difference between \cite{imgcm} and this paper. 
\par
Our work is not the first to take the document type into account. Many click models are proposed for \textit{aggregated search}, as shown in \textit{federated click model} \cite{FCM} and \textit{vertical click model} \cite{VCM}. But the model introduced in this paper does not aim to study \textit{aggregated search}. In fact, in \textit{aggregated search}, candidate results are from multiple sources and have different types. \cite{FCM} and \cite{VCM} model such heterogeneity. However, we pay attention to a specific and rarely mentioned type, which is image-oriented document. 
\par
Existing click models are mainly based on \textit{probabilistic graphical model} (PGM) framework. Two kinds of user behaviors, \textit{a document is examined by a user} and \textit{a document attracts a user}, are represented as hidden events. In existing click models, while the probability of the former is modeled differently, the latter is usually modeled by a parameter with a query and a document as its indicator, and the parameters are usually estimated by generative maximum likelihood under specific assumptions. Many different click models have been developed in the past, among which PBM and CM are two classical ones. UBM extends the hypothesis of PBM by introducing a new examination parameter. DBN and CCM extend CM to model the probability for a document to get examined depending on the previous documents. This paper chooses PBM and UBM to extend them by incorporating \textit{vision bias}, but any model with an examination bias can be extended in a similar way. Feature vectors are used to represent a parameter, \textit{vision bias}, and we estimate its value by regression following \cite{trust}, \cite{google} and \cite{RegEM}. This approach handles the data sparsity problem better, and also contributes to the extensibility of our work. 
\par
In recent years, other approaches taking advantage of dense vectors of context attributes to estimate biases in click models are proposed, such as \cite{encoder} and \cite{neural}. These works use complex deep neural networks to build up click models and encode the context attributes by vectors. Compared with our model, the models in \cite{encoder} and \cite{neural} demand much time for finding a solution and consume a large amount of computing resources.
\par
As is mentioned above, the parameter inference method in this paper follows \cite{trust}, \cite{google} and \cite{RegEM}. Since neural network processes dense vectors more efficient than Gradient Boosted Decision Tree (GBDT)\cite{GBDT}, regression-based EM altorithm in this paper uses a simple neural network (NN) as its predictor. Because we assume that image features have impacts on the estimation of examination probabilities, regression-based EM algorithm is taken as an approach to infer the examination probability instead of relevance, which is different from \cite{trust} 
and \cite{RegEM}. Particularly, \cite{google} assumes that context attributes influence the estimation of examination probability. But our paper differs from \cite{google} in two aspects. First, the image-oriented search engine is an independent and specific case of estimating biases by incorporating context attributes. Second, \cite{google} continues to use probabilistic graph structure in previous classical click models such as PBM and UBM, and incorporate any bias into the model by making examination probability dependent of different context attributes. However, our method explicitly introduces a new structure for the probabilistic graph. This explicit modeling approach makes the model more expressive. 

%% file: modeling.tex
This section states the \textit{vision bias}, and derives the expressions of the probabilities for relevant hidden variable events. Then we introduce the methods to infer the parameters proposed in our model.
\par
Before introducing any specific click model, we first declare definitions and notations used in this paper. In any search session $s$ where a query $q$ is given by the user and the documents are displayed, a document at rank $r$ is denoted by $d_{r}$. We define four binary variables, $C_{r}$, $E_{r}$, $\widetilde{E}_{r}$ and $R_{r}$, as the user click, user examination based on position, real user examination and document relevance event at rank $r$. Specifically, $C_{r} = 1$ when a user clicks the document at rank $r$, $E_{r} = 1$ when any document gets examined at rank $r$, $\widetilde{E}_{r} = 1$ when a certain document at rank $r$ is  examined, and $R_{r} = 1$ when the document at rank $r$ is relevant to the query $q$. 
\par
In this paper, we use PBM and UBM as baseline models and apply \textit{vision bias} into their assumptions to extend them. 

\begin{figure}[H]
\subfigure[Examination Hypothesis]{
    \label{Fig.sub.eh}
    \includegraphics[width=0.18\textwidth]{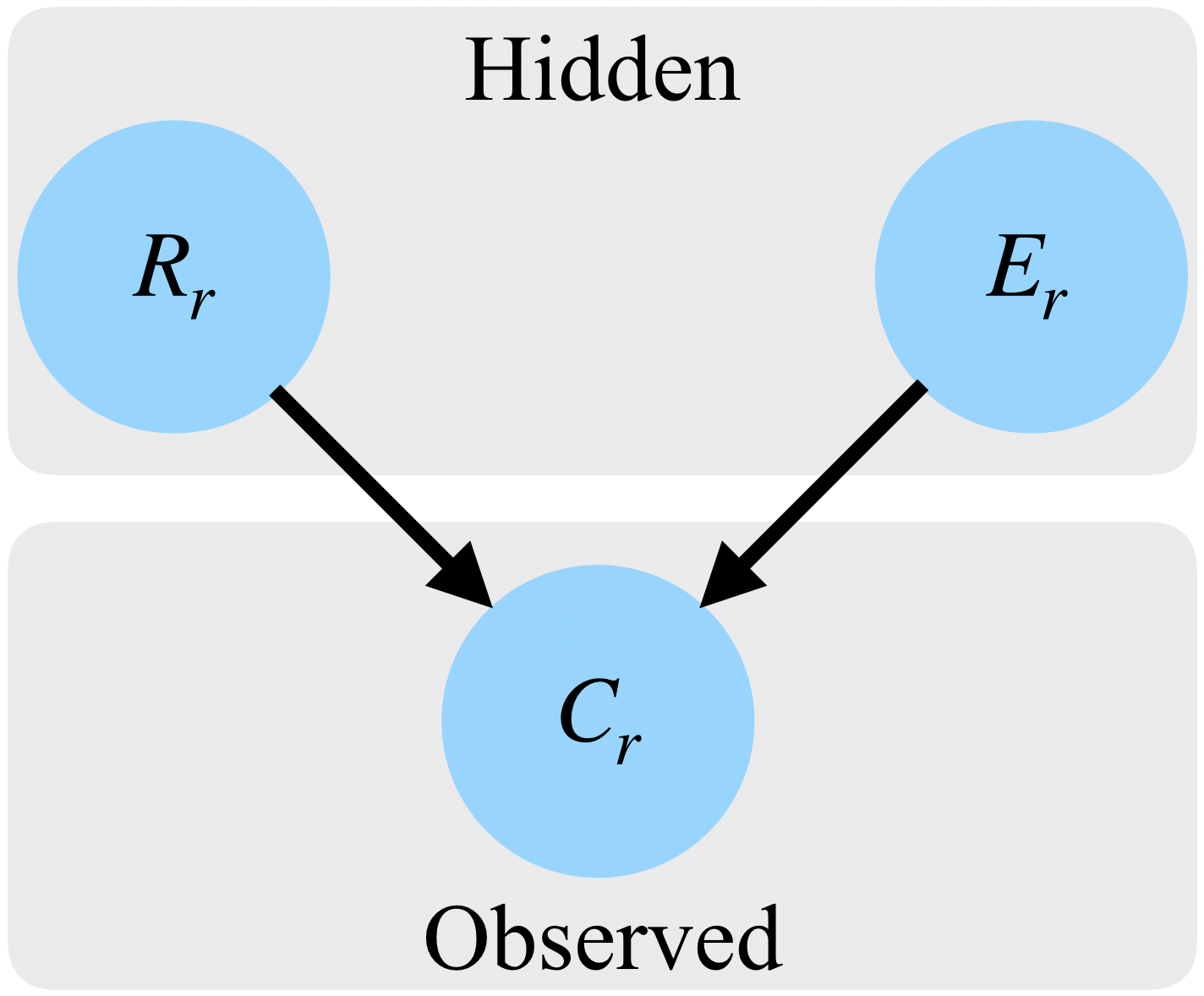}
}
\subfigure[Vision Bias Hypothesis]{
    \label{Fig.sub.vbh}
    \includegraphics[width=0.18\textwidth]{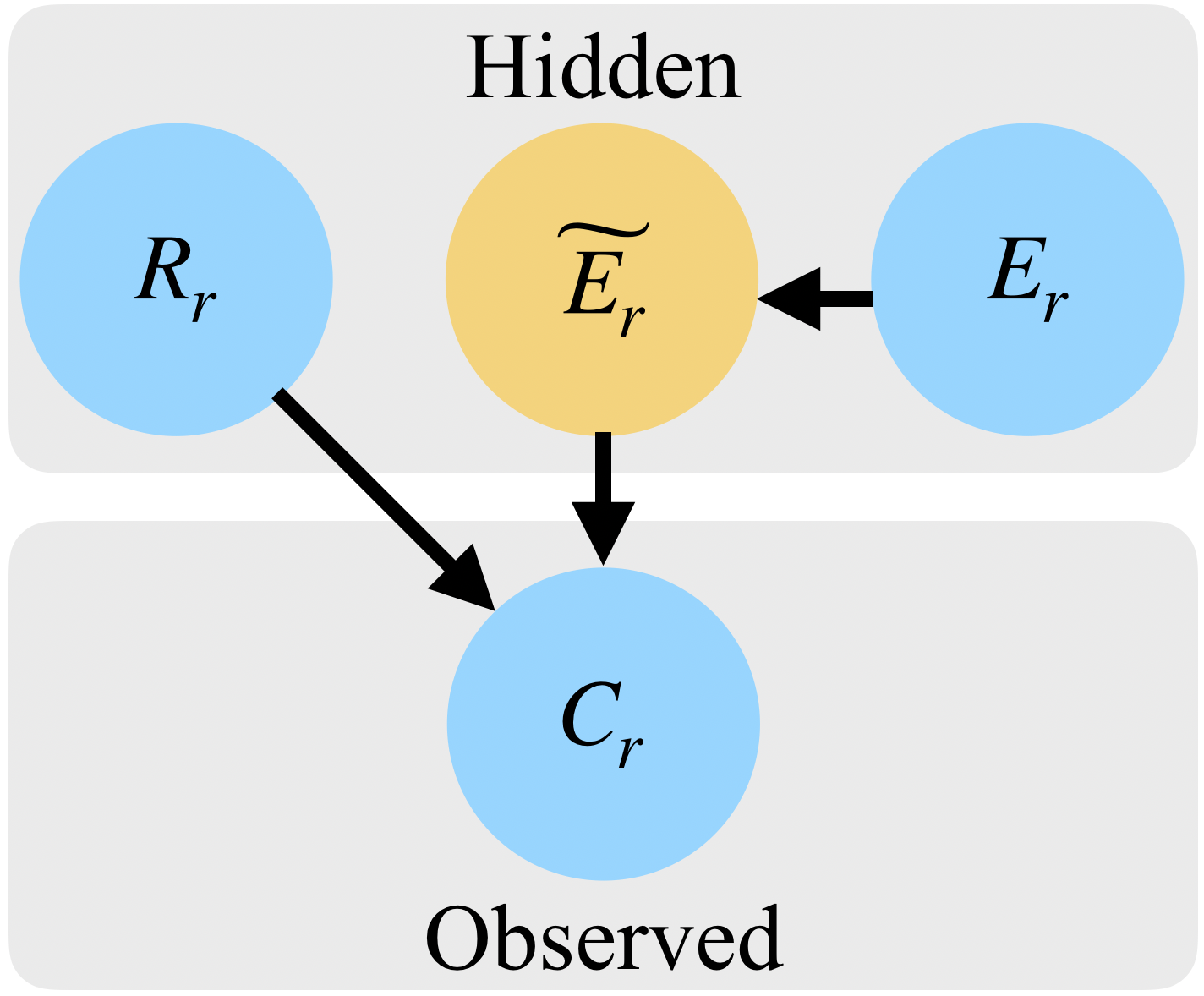}
}
\caption{The graphical models of the examination hypothesis and the \textit{vision bias} hypothesis}
\label{Fig.main}
\end{figure}

\subsection{Vision Bias}

When \textit{vision bias} is applied to PBM and UBM, the proposed hypothesis partially preserves the original hypothesis, but adds \textit{vision bias} to it as a new part. The proposed hypothesis assumes that the real examination probability of a given document in the search session not only depends on its position or the position of its previously clicked document, but also is under the influence of its visual features. Besides, a document is clicked if and only if it is examined by the user and recognized as relevant. As in original PBM and UBM, the relevance of the given document is uniquely affected by the query and the content of the document.
\par
To further describe how \textit{vision bias} works together with the original position-based examination bias, we conclude the extended model by four hypotheses as below:
\par
\begin{itemize}
\item A document (image-oriented) is clicked if and only if it is examined and relevant.
\item Position of the document influences its examination probability.
\item Even if an document is not placed in a position where documents tend to be examined, eye-catching visual features can increase its examination probability.
\item The relevance only depends on the query and the document.
\end{itemize}
Figure \ref{Fig.sub.eh} shows the structure of PBM and UBM. In comparison, Figure \ref{Fig.sub.vbh} illustrates the hypothesis of \textit{vision bias}. The figure shows that a latent event $\widetilde{E}$ is inserted between the click event $C$ and the event $E$. This event is an intermediate state where the document is examined. It represents the result for a document to be examined, and depends on the document position and its visual features. Though in the figure, it is represented by $\widetilde{E}_r$, in fact it not only depends on the rank of the document. Here, $r$ only means that $\widetilde{E}$ corresponds to the document at rank $r$.
\par
In Figure \ref{Fig.main}, each circle represents a certain event, and the edge connecting any event describes the dependency between them. Each document is placed in a certain position in the layout of the image-oriented search engine. When they are displayed, a thumbnail is shown so that their visual features is directly exposed to the users. Each user comes to the search engine with a formulated query, and the search engine provides a list of documents. The possibility for a document to be examined is different when it is placed on different positions. If the position where it is placed tends to be examined by the user, then it is more likely that the image placed there get examined. However, different documents placed on the same position still have different opportunities to be examined. If an image-oriented document is not placed on a position that tends to be examined, but has eye-catching visual features like large size or bright colors, still it is more likely for this document to be examined. In Figure \ref{Fig.sub.vbh}, the event that a document is examined is divided into two stages: 
First, only how position influences the examination probability for the document is considered. Second, we also take image visual features into consideration to obtain its real examination probability.
When the users examine a document, they decide whether it is relevant to the query. If an examined document is relevant, then the user will click the document. 
\par
The examination bias in PBM and UBM measures the possibility of examination based on position, while \textit{vision bias} measures the examination probability under the influence of any factor determined by the visual features. In fact, PBM and UBM can be regarded as special cases of the extended model. If the effects of visual features on the image-oriented documents' examination probability are ignored, in other words, \textit{vision bias} is ignored, and we only consider the examination probability as an event only affected by position, then the examination in our extended model is equivalent to the original \textit{examination hypothesis}.
\subsection{Incorporation}
The switch from standard PBM and UBM to the extended models is quite simple. The details of PBM and UBM can be found in \cite{PBM} and \cite{UBM}, but in this paper we will not discuss about it. This section only takes UBM as an example to introduce how to extend a baseline model. The method to incorporate \textit{vision bias} into PBM is similar. For UBM, we only need to replace the original model assumptions with the following ones without changing any other specifications. We use \textbf{$C_{<r_{d}}$} to represent all the click events previous to rank $r$, and use \textbf{C} to represent all the click events in the search session. $\gamma_{rr'}$ is the position bias, and $\sigma_{d}$ is the \textit{vision bias}. $\widetilde{E}_{r}$ and $E_{r}$ in the equations are values corresponding to the document at rank $r$, but the values of $E$ and $\widetilde{E}$ should not only depend on the rank of the current document. Instead, because the conditional probability in Equation \ref{cplt} depends on the document itself, $\widetilde{E}$ should also depend on a specific document.
\par
\begin{equation}
P\left(E_{r} = 1 \mid \textbf{$C_{<r_{d}}$}\right) = \gamma_{rr'}
\end{equation}
\begin{equation}
    P\left(\widetilde{E}_{r} = 1 \mid E_{r} = 0\right) = \sigma_{d}
    \label{cplt}
\end{equation}
\begin{equation}
    P\left(\widetilde{E}_{r}=1 \mid E_{r}=1\right) = 1
\end{equation}
If $\sigma_{d}$ is alway set to 0, the event $\widetilde{E}_{r}$ is equivalent to the event $E_{r}$. In other words, the effects of visual features on the examination probability are ignored, and we consider an image-oriented document's examination probability only affected by the position. In this case, the extended model degrades into a standard one.
\par
The extended UBM model has
\begin{equation}
\begin{aligned}
P\left(\widetilde{E}_{r} = 1 \mid \textbf{C}\right) &=P\left(E_{r} = 1 \mid \textbf{$C_{<r_{d}}$}\right) \cdot P\left(\widetilde{E}_{r} = 1 \mid E_{r} = 1 \right)\\
 &+ P\left(E_{r} = 0 \mid \textbf{$C_{<r_{d}}$}\right) \cdot P\left(\widetilde{E}_{r} = 1 \mid E_{r} = 0 \right)\\
&= P\left(E_{r} = 1 \mid \textbf{$C_{<r_{d}}$}\right) \\
&+ \left(1 - P\left(E_{r} = 0 \mid \textbf{$C_{<r_{d}}$}\right)\right) \cdot P\left(\widetilde{E}_{r} = 1 \mid E_{r} = 0 \right)\\
&= \gamma_{_{rr'}} + \left(1 - \gamma_{_{rr'}}\right)\sigma_{d}
\end{aligned}
\label{hypo}
\end{equation}
Thus, in the extended UBM model, the conditional probability given the clicks of previous documents for a click event of document $d$ at rank $r$ is 
\begin{equation}
\begin{aligned}
P\left(C_{r} = 1 \mid \textbf{C}\right) &= P\left(\widetilde{E}_{r} = 1 \mid \textbf{$C_{<r_{d}}$}\right) \cdot P\left(R_{r} = 1\right)\\
&= \alpha_{_{qd}}\left(\gamma_{_{rr'}} + \left(1 - \gamma_{_{rr'}}\right)\sigma_{d}\right)
\end{aligned}
\end{equation}
Here, the relevance between query and document is denoted as $\alpha_{qd}$. The probability of a click event at rank $r$ is given as 
\begin{equation}
\begin{aligned}
\sum_{j=0}^{r_{d}-1}  P\left(C_{j}=1\right)  \cdot \left( \prod_{k=j+1}^{r_{d}-1}\left(1 - \alpha_{_{qd_{k}}}\left(\gamma_{_{kj}} + \left(1 - \gamma_{_{kj}}\right)\sigma_{d_{k}}\right)\right) \right) \cdot \\
\alpha_{_{qd}}\left(\gamma_{_{r_{d}j}} + \left(1 - \gamma_{_{r_{d}j}}\right)\sigma_{d}\right) 
\end{aligned}
\end{equation}
\begin{equation}
    P\left(C_{0} = 1\right) = 1
\end{equation}
The parameters of the extended UBM model can be estimated via EM algorithm using click-through logs. Thus, the log-likelihood of generating log data $\mathcal{D}$ is given as
\begin{equation}
\begin{aligned}
\log P\left( \mathcal{D}\right) &= \sum_{\left(q,d,r,c \in \mathcal{D}\right)} \left[ c \cdot \log \left(\alpha_{_{qd}}\left(\gamma_{_{rr'}} + \left(1 - \gamma_{_{rr'}}\right)\sigma_{d}\right)\right)
\right.
\\
&+ \left. \left(1-c\right) \cdot \log\left(1 - \alpha_{_{qd}}\left(\gamma_{_{rr'}} + \left(1 - \gamma_{_{rr'}}\right)\sigma_{d}\right)\right) \right]
\label{ll}
\end{aligned}
\end{equation}

 \begin{displaymath}
 r' = max \{k \in \{ 0, ..., r-1 \}:  c_{k} = 1\}
\end{displaymath}
where $c$ denotes whether a document is clicked.

%% file: infer.tex
In EM algorithm, parameters are estimated by alternatively carrying out Expectation and Maximization steps (E-steps and M-steps) until the estimations converge. Parameters at iteration $t$ are estimated based on the results at iteration $t-1$. 
\subsection{Standard EM Algorithm}
In the Expectation step of standard EM algorithm, the goal is to maximize the log-likelihood of the model if the click-through logs are given. In order to simplify the computation, as is shown in \cite{DLR77}, we derive the posterior distribution of the hidden variables representing the events in the model, and aim to find a value for any parameter to optimize this probability in the M-step.
\par
Due to the space limitation, this paper directly give iterative formulas for parameters in the extended UBM model as follows.

\begin{equation}
\begin{aligned}
\alpha_{_{qd}}^{t+1} = \frac{\sum_{s \in S_{q}} \left(c_{d}+\left(1-c_{d}\right)\alpha_{qd}^{t}\frac{1-\left(\gamma_{rr'}^{t}+\sigma_{d}^{t}-\gamma_{rr'}^{t}\sigma_{d}^{t}\right)}{1-\alpha_{qd}^{t}\left(\gamma_{rr'}^{t}+\left(1-\gamma_{rr'}^{t}\right)\sigma_{d}^{t}\right)} \right)}{\sum_{s \in S_{q}} 1}
\end{aligned}
\end{equation}
\begin{equation}
\begin{aligned}
\gamma_{rr'}^{t+1}=\frac{\sum_{s \in S_{q}}\left(c_{d}\frac{\gamma_{rr'}^{t}}{\gamma_{rr'}^{t}+\sigma_{d}^{t}\left(1-\gamma_{rr'}^{t}\right)}+\left(1-c_{d}\right)\frac{\gamma_{rr'}^{t}\left(1-\alpha_{qd}^{t}\right)}{1-\alpha_{qd}^{t}\left(\gamma_{rr'}^{t}+\left(1-\gamma_{rr'}^{t}\right)\sigma_{d}^{t}\right)}\right)}{\sum_{s \in S_{q}} 1}
\end{aligned}
\end{equation}
\begin{equation}
\begin{aligned}
\sigma_{d}^{t+1}=\frac{\sum_{s \in S_{q}}\left(c_{d}\frac{\sigma_{d}^{t}\left(1-\gamma_{rr'}^{t}\right)}{\gamma_{rr'}^{t}+\left(1-\gamma_{rr'}^{t}\right)\sigma_{d}^{t}}+\left(1-c_{d}\right)\frac{\sigma_{d}^{t}\left(1-\gamma_{rr'}^{t}\right)\left(1-\alpha_{qd}^{t}\right)}{1-\alpha_{qd}^{t}\left(\gamma_{rr'}^{t}+\left(1-\gamma_{rr'}^{t}\right)\sigma_{d}^{t}\right)}\right)}{\sum_{s \in S_{q}}\left(c_{d}\frac{\sigma_{d}^{t}\left(1-\gamma_{rr'}^{t}\right)}{\gamma_{rr'}^{t} + \left(1-\gamma_{rr'}^{t}\right)\sigma_{d}^{t}}+\left(1-c_{d}\right)\frac{\left(1-\gamma_{rr'}^{t}\right)\left(1-\sigma_{d}^{t}\alpha_{qd}^{t}\right)}{1-\alpha_{qd}^{t}\left(\gamma_{rr'}^{t}+\left(1-\gamma_{rr'}^{t}\right)\sigma_{d}^{t}\right)}\right)}
\end{aligned}
\end{equation}
For the extended PBM model, the formulas are given as
\begin{equation}
\begin{aligned}
\alpha_{_{qd}}^{t+1} = \frac{\sum_{s \in S_{q}} \left(c_{d}+\left(1-c_{d}\right)\alpha_{qd}^{t}\frac{1-\left(\gamma_{r}^{t}+\sigma_{d}^{t}-\gamma_{r}^{t}\sigma_{d}^{t}\right)}{1-\alpha_{qd}^{t}\left(\gamma_{r}^{t}+\left(1-\gamma_{r}^{t}\right)\sigma_{d}^{t}\right)} \right)}{\sum_{s \in S_{q}} 1}
\end{aligned}
\end{equation}
\begin{equation}
\begin{aligned}
\gamma_{r}^{t+1}=\frac{\sum_{s \in S_{q}}\left(c_{d}\frac{\gamma_{r}^{t}}{\gamma_{r}^{t}+\sigma_{d}^{t}\left(1-\gamma_{r}^{t}\right)}+\left(1-c_{d}\right)\frac{\gamma_{r}^{t}\left(1-\alpha_{qd}^{t}\right)}{1-\alpha_{qd}^{t}\left(\gamma_{r}^{t}+\left(1-\gamma_{r}^{t}\right)\sigma_{d}^{t}\right)}\right)}{\sum_{s \in S_{q}} 1}
\end{aligned}
\end{equation}
\begin{equation}
\begin{aligned}
\sigma_{d}^{t+1}=\frac{\sum_{s \in S_{q}}\left(c_{d}\frac{\sigma_{d}^{t}\left(1-\gamma_{r}^{t}\right)}{\gamma_{r}^{t}+\left(1-\gamma_{r}^{t}\right)\sigma_{d}^{t}}+\left(1-c_{d}\right)\frac{\sigma_{d}^{t}\left(1-\gamma_{r}^{t}\right)\left(1-\alpha_{qd}^{t}\right)}{1-\alpha_{qd}^{t}\left(\gamma_{r}^{t}+\left(1-\gamma_{r}^{t}\right)\sigma_{d}^{t}\right)}\right)}{\sum_{s \in S_{q}}\left(c_{d}\frac{\sigma_{d}^{t}\left(1-\gamma_{r}^{t}\right)}{\gamma_{r}^{t} + \left(1-\gamma_{r}^{t}\right)\sigma_{d}^{t}}+\left(1-c_{d}\right)\frac{\left(1-\gamma_{r}^{t}\right)\left(1-\sigma_{d}^{t}\alpha_{qd}^{t}\right)}{1-\alpha_{qd}^{t}\left(\gamma_{r}^{t}+\left(1-\gamma_{r}^{t}\right)\sigma_{d}^{t}\right)}\right)}
\end{aligned}
\end{equation}
\subsection{Regression-based EM Algorithm}
In order to capture more sophisticated information contained in the image-oriented documents, specifically the visual features of images, regression-based EM algorithm is applied to infer the parameters. Regression-based EM algorithm is proposed and developed in \cite{trust}, \cite{google} and \cite{RegEM}. 
In our extended model, regression-based EM algorithm is used in the estimation of the parameter $\sigma_{d}$. 
It does not change the E-step in standard EM algorithm. 
The goal of its M-step is to find a proper function $f$ that reflects the dependency between $\textbf{x}_{d}$ and $\sigma_{d}^{t+1}$, and thus to maximize the likelihood inferred from the E-step.
\par
Regression-based EM algorithm is preferred in the extended model for several reasons: (1) User behaviors are sparse. In the training data generated from click logs, some documents appear for many times, while others might only appear for a few times. (2) To incorporate the visual features of documents, the identifiers of documents used in standard EM algorithm have to be replaced by the feature vectors of documents. 
Regression-based EM algorithm is used in the extended model to overcome these difficulties. It extracts a feature vector $\textbf{x}$ from every image-oriented document to implement the M-step, and incorporates the visual features of image-oriented documents in the estimation of the parameter $\sigma_{d}$.
\par
Regression-based EM algorithm implements the function $f$ via regression, and fits the neural network (NN) to regress the visual features $\textbf{x}_{d}$ to the derived target value $P \left( \widetilde{E}_{r}=1 \mid E_{r}=0 \right)$. As in \cite{trust}, \cite{google} and \cite{RegEM}, we convert the regression problem to a classification problem and specific reasons for converting is beyond the scope of this paper. 
The structure of NN classifier is fairly concise. It is composed of multilayer perceptron (MLP)  
with \textit{sigmoid} activation function 
, and \textit{softmax} 
output layer \cite{goodfellow2016deep}. When implementing regression-based EM algorithm in the estimation of \textit{vision bias}, we sample $l$ based on the value of $\sigma_{d}$ given in the M-step in standard EM algorithm with an average threshold to determine positive and negative samples. After NN classifier is trained, predicted probabilities for a sample to be in the positive class will be the estimation of the parameter $\sigma_{d}$ corresponding to it. 
Algorithm \ref{Alg.RegEM} summarizes the complete process of regression-based EM algorithm for the extended model.

\begin{algorithm}
\caption{Regression-based EM algorithm for the extended model.}
\label{Alg.RegEM}
\begin{algorithmic}[1]
\Require
$S$.
\Ensure
$f\left( \textbf{x} \right)$, $\gamma_{rr'}$, and $\alpha_{qd}$.
\State Extract visual features $\textbf{x}_{d}$.
\State Set up $\gamma_{rr'}$, $\alpha_{qd}$ and $\sigma_{d}$.
\Repeat
\For{each $s \in S$}
\State Estimate \textit{vision bias} and update $\alpha_{qd}$ and $\gamma_{rr'}$.
\label{code:fram:prob}
\EndFor
\State Sample $l \in \left(0, 1\right)$ from $\sigma_d$ to form training dataset $\mathcal{D}$.
\State $F\left( \textbf{x} \right) = MLP\left(F\left( \textbf{x} \right), \mathcal{D}\right)$
\State $f\left( \textbf{x} \right) = \textit{softmax}\left(F\left( \textbf{x} \right)\right)$
\State $\sigma_{d} = f\left( \textbf{x}_{d} \right)$
\Until{Convergence.}\\
\Return $f\left(\textbf{x}\right)$, $\gamma_{rr'}$, and $\alpha_{qd}$.
\end{algorithmic}
\end{algorithm}
\par
The 64-dimensional visual features are extracted from the image-oriented documents.
The feature is 64-dimensional, which have enough capacity to represent the visual information of a document. The direct output of ResNet \cite{ResNet} has high dimension, which is redundant and hard to be optimized. Therefore, we pretrain a neural network with user click behavior to classify whether the image was clicked or not, and treat the image-oriented document as an input image and emplot ResNet-50 network which is widely used for image representation. After coming through all layers in ResNet, an input image becomes a 2048-dimensional feature. Then a fully connected layer is employed to reduce the redundant information in the past feature. Finally, the 64-dimensional features are extracted as the visual feature that we adopt for regression-based EM algorithm.

%% file: exp.tex
\begin{figure} 
  \centering
  \includegraphics[width=0.35\textwidth]{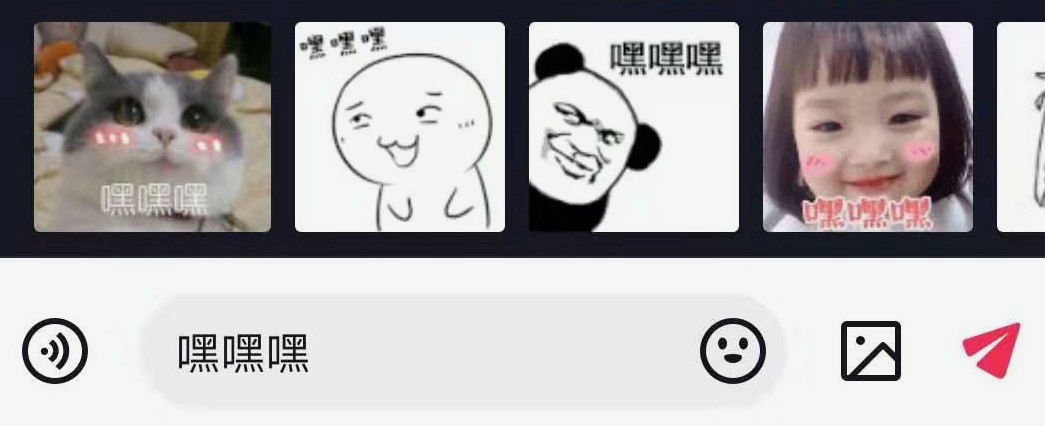}
  \caption{The image-oriented search engine with a Chinese query (\textit{hey hey hey}).}
  \Description{The image-oriented search engine with a Chinese query (\textit{hey hey hey}).}
  \label{Fig.tiktok2}
\end{figure}

In this section, we evaluate the performance of different models on the dataset collected from an online image-oriented search engine in \textit{Douyin} (\href{www.douyin.com}{www.douyin.com}, 
a video community app from \textit{ByteDance}). To the best of our knowledge, no public datasets are available for the experiments in this paper, and \textit{Douyin} dataset is the only dataset that we can use. This dataset is generated from click-through logs searching for memes as shown in Figure \ref{Fig.tiktok2}. Images (memes) appear on the bottom of the screen horizontally. When users are chatting, they can type in queries to search images to send. The number of examined images in each search session is not fixed, since users can swipe the screen to the left or to the right to browse different images. The dataset does not involve any private user information. Based on our knowledge, there are no other public datasets with both click-through logs and original images from an image-oriented search engine. Thus, the experiments are only performed on this dataset. 

\subsection{Experiment Settings}

The dataset is created based on the raw click-through logs. The raw logs have 3,460,178 records, each for a search session containing the clicked image, all the displayed images in that search session and the corresponding query. In order to obtain the dataset, we first filter the queries by standardizing synonyms and removing uncommon characters. Then the dataset is splited into two parts as the training dataset and the test dataset. The first 75\% search sessions are the training dataset and the last 25\% are the test dataset. The generation of test dataset requires dropping queries and documents that do not appear in the training dataset from the test dataset, following the experiments in \cite{log}. In the end, the dataset contains 3,251,801 search sessions randomly sampled from the click-through logs. In the training dataset, there are 2,438,434 search sessions, 128,525 unique queries and 184,335 unique images. In the test dataset, there are 813,367 search sessions, 57,287 unique queries and 141,546 unique images. 
We pay attention to the frequency of queries, and summarize it in Table \ref{freq_q}. 

\begin{table}
  \caption{Query Frequency Statistics}
  \label{freq_q}
  \begin{tabular}{p{2.5cm}p{1.5cm}p{1.5cm}p{1.5cm}}
    \toprule
    Query frequency&Training dataset&Test dataset\\
    \midrule
    1-10&88,875&39,289\\
    10-30&25,080&11,224\\
    30-100&9,034&4,259\\
    100-500&3,871&1,801\\
    500-2000&1,104&493\\
    2000-10000&404&159\\
    $\geq$ 10000&157&62\\
  \bottomrule
\end{tabular}
\end{table}


\par
Following \cite{neural}, \cite{noise}, and \cite{trust}, we compare the extended click models with their baseline models without considering models beyond the scope of click models or other biases. The initial value $\alpha$ and $\gamma$ in UBM are set to 0.2 and 0.5 as is recommended in \cite{UBM}. They are set to 0.5 and 0.5 in PBM. 
In the neural network used in regression-based EM algorithm, the learning rate is set to 0.05. The multilayer perceptron (MLP) \cite{goodfellow2016deep} consists of two hidden layers with 16 and 8 perceptrons separately.
The whole framework is implemented via \textit{PySpark} and any other parameters can be found in default setting of \textit{MultilayerPerceptronClassifier} in \textit{PySpark}.
In this section, we define vUBM-2 to represent the extended UBM model, and vPBM-2 to represent the extended PBM model. 
\par
To discuss about another possible Bayesian relationship between $\widetilde{E}_{r}$ and $E$ in \ref{Fig.sub.vbh}, here two additional models, vUBM-1 and vPBM-1, are built up for comparison. The proposal of these two models is inspired by \cite{intent}, where the author proposes a new bias and makes it multiplied by the relevance. In a similar way, vUBM-1 and vPBM-1 model \textit{vision bias} as $P\left(\widetilde{E}_{r} = 1 \mid E_{r} = 1 \right)$, and make it multipled by the original position bias. To obtain the probabilistic graph structure for these two models, we only need to replace Equation \ref{hypo} with Equation \ref{yijie}. When we use Equation \ref{yijie}, the model should be re-trained and inferred by the new formula to keep the comparison fair.
\begin{equation}
\begin{aligned}
P\left(\widetilde{E}_{r} = 1 \mid \textbf{C}\right) &=P\left(E_{r} = 1 \mid \textbf{$C_{<r_{d}}$}\right) \cdot P\left(\widetilde{E}_{r} = 1 \mid E_{r} = 1 \right)\\
&= \gamma_{_{rr'}}{\sigma}_{d}
\end{aligned}
\label{yijie}
\end{equation}

\subsection{Click Prediction}
This section evaluates how models perform in click prediction tasks. Several different metrics can be used in the evaluation. Modeling \textit{Vision bias}, the inherent bias that this paper studies, helps improve the performance of the extended models.
\subsubsection{Log-likelihood}

Log-likelihood is a widely-used metric to measure the performance of click models in click predictions. Standard log-likelihood is computed as follows.
\begin{displaymath}
LL\left(\mathcal{D}\right)=\sum_{s \in S} \sum_{r=1}^{n}\log P\left(C_{r}=c_{r}^{s} \mid C_{<r}=c_{<r}^{s}\right) 
\end{displaymath}
Each search session is represented as $s$ and there are $n$ documents in a session. This metric is always non-positive. The higher the log-likelihood is, the better the click model performs. A perfect click model will have a log-likelihood equal to 0. 
\par
The results of the experiment are shown in Table \ref{ll2}. The numbers in this table for vUBM-2 and vPBM-2 have passed a \textit{paired t-test} (p<0.01). Both vUBM-2 and vPBM-2 significantly outperform baselines, which means better performance in click prediction. The results for comparison models vPBM-1 and vUBM-1 are also shown in Table \ref{ll2}.


\begin{table}
  \caption{Performance on log-likelihood. Extended models vPBM-2 and vUBM-2 always have the greatest improvements.}
  \label{ll2}
  \begin{tabular}{ccc}
    \toprule
    &Log-likelihood&Improvement\\
    \midrule
    UBM&-0.433&-\\
    vUBM-1&-0.431&+0.46\%\\
    vUBM-2&-0.409&+\textbf{5.54}\%\\
    \hline
    PBM&-0.429&-\\
    vPBM-1&-0.426&+0.70\%\\
    vPBM-2&-0.409&+\textbf{4.66}\%\\
  \bottomrule
\end{tabular}
\end{table}
\subsubsection{Perplexity}


Perplexity is another widely-used metric to measure the data fitness of click models. 
Perplexity is computed for click events at each rank of a search session. The perplexity of the whole test dataset is the average of perplexities at different ranks. The formula to compute perplexity at rank $r$ is given as
\begin{displaymath}
p_{r}=2^{-\frac{1}{|S|}\sum_{s \in S}\left(c_{r}^{s}\log_{2}\overline{q}_{r}^{s}+\left(1-c_{r}^{s}\right)\log_{2}\left(1-\overline{q}_{r}^{s}\right)\right)}
\end{displaymath}
where each search session is represented as $s$ and $\overline{q}_{r}^{s}$ is the probability for a user to click the document at rank $r$ in the search session $s$. A smaller perplexity indicates better data fitness, and a perfect click model will have a perplexity of $1$. The improvement of perplexity value $p_{1}$ over $p_{2}$ is calculated as $\frac{p_{2}-p_{1}}{p_{2}-1} \times 100\%$.
\par
The total perplexity of different models are reported in Table \ref{perp}. Since in the dataset, a query corresponds to an unfixed number of images and a few query can have a very large number of candidate images, when computing the total perplexity, only perplexities over the first 10 ranks are averaged. It is observed that both vUBM-2 and vPBM-2 significantly outperform baselines, which indicates better data fitness. The results for comparison models vPBM-1 and vUBM-1 in perplexity are also shown in Table \ref{perp}.
\begin{table}
  \caption{Performance on total perplexity. Extended models vPBM-2 and vUBM-2 have the lowest perplexity and the greatest improvements.}
  \label{perp}
  \begin{tabular}{ccc}
    \toprule
    &Perplexity&Improvement\\
    \midrule
    UBM&1.417&-\\
    vUBM-1&1.409&+1.92\%\\
    vUBM-2&1.388&+\textbf{6.95}\%\\
    \hline
    PBM&1.412&-\\
    vPBM-1&1.402&+2.43\%\\
    vPBM-2&1.381&+\textbf{7.52}\%\\
  \bottomrule
\end{tabular}
\end{table}
\subsubsection{Comparison}
This section discusses the necessity of Equation \ref{hypo} compared with modeling \textit{vision bias} by Equation \ref{yijie}. As is mentioned in the last section, this is to demonstrate the rationality of the probabilistic graph structure in the extended models.
\par
The experiment results can be found in Table \ref{ll2} and \ref{perp}. From these tables it can be found that the extended models outperform the two compared models, vUBM-1 and vPBM-1, both in log-likelihood and perplexity. It shows that in the click prediction task, our choice of Equation \ref{hypo} is superior to Equation \ref{yijie} to model \textit{vision bias}. Thus, the following sections only use Equation \ref{hypo} as a modeling foundation.
\subsection{Sparsity Handling}
This section discusses about how extended models contribute to sparsity handling. In user click logs, queries and documents are sparse. The parameters in click models are tricky to be accurately estimated since the sparsity. 
We independently experiment on the situations where extended models deal with sparse training data, specifically, low-frequency queries and higher rank positions, and the results show that our proposed models have superior performance on the sparse user data.
\subsubsection{Data fitness}
We first use log-likelihood to evaluate the extended model by computing it separately for different query frequencies. Table \ref{loglikeli} shows the log-likelihood of different models for queries appearing in the training dataset for less than 100 times. Both extended models achieve better performance than their baselines in these low-frequency queries. 
In this table, the lower the query frequency is, the greater improvements the extended models achieve, and the greatest improvements are both achieved on the least frequent queries. Thus, on low-frequency queries, the extended models perform well in log-likelihood. 
\begin{table}
  \caption{Performance on log-likelihood for different query frequencies and corresponding improvements. Extended models achieve the greatest improvements on the most sparse training data. }
  \label{loglikeli}
  \begin{tabular}{c|cc|cc}
    \hline
    Freq.&UBM&vUBM-2&PBM&vPBM-2\\
    \hline
    1-10&-0.447&\textbf{-0.296 (+33.78\%)}&-0.444&\textbf{-0.295 (+33.56\%)}\\
    10-30&-0.434&-0.352 (+18.89\%)&-0.429&-0.351 (+18.18\%)\\
    30-100&-0.426&-0.400 (+6.10\%)&-0.424&-0.401 (+5.42\%)\\
  \hline
\end{tabular}
\end{table}
\par
\begin{figure} 
  \centering
  \includegraphics[width=0.45\textwidth]{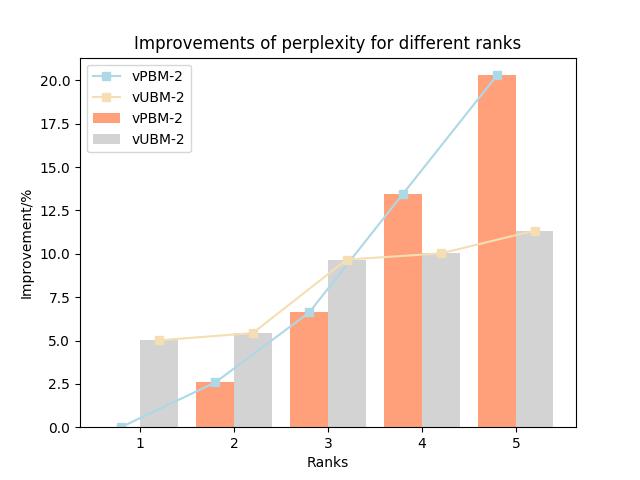}
  \caption{Improvements of perplexity for different ranks. With the rank becoming higher, the training data for that rank becomes more sparse. In this figure, extended models achieve significant improvements in sparse training data.}
  \Description{Improvements of perplexity for different ranks. With the rank becoming higher, the training data for that rank becomes more sparse. In this figure, extended models achieve more improvements in sparse training data.}
  \label{Fig.pp_r}
\end{figure}

Following \cite{UBM} and \cite{CCM}, we use perplexity to evaluate the extended model, and compute it separately for different ranks and different query frequencies. When computing perplexity for different ranks, because on average each query corresponds to 5 documents, figures and tables in this section only show the results from rank 1 to rank 5. In Table \ref{pp_rank}, the extended models achieve the best perplexity in the highest rank. Figure \ref{Fig.pp_r} shows the improvement 
trend over different ranks. With the rank raising up, the improvements also increase. Because in the dataset, not all search sessions have the same number of documents, higher rank means less training data. Thus, with higher rank and more sparse training data, extended models performs better than their baselines in perplexity, and indicates better capability to handle sparsity.
\begin{table}
  \caption{Performance on perplexity for different ranks and corresponding improvements. The higher the rank, the more sparse the training data. Both extended models achieve the greatest improvements on the highest rank, which corresponds to the most sparse training data.}
  \label{pp_rank}
  \begin{tabular}{c|cc|cc}
    \hline
    Rank&UBM&vUBM-2&PBM&vPBM-2\\
    \hline
    1&1.701&1.666 (+4.99\%)&1.708&1.708 (+0.00\%)\\
    2&1.609&1.577 (+5.25\%)&1.608&1.592 (+2.63\%)\\
    3&1.531&1.480 (+9.60\%)&1.508&1.475 (+6.50\%)\\
    4&1.413&1.372 (+9.93\%)&1.399&1.345 (+13.53\%)\\
    5&1.331&\textbf{1.293 (+11.48\%)}&1.318&\textbf{1.254 (+20.13\%)}\\
  \hline
\end{tabular}
\end{table}
\par
To further compare the extended models, we divide the search sessions in the test dataset into several parts based on how many times the query appears in the training dataset. Similar to the analysis of log-likelihood, we report the perplexity results for queries appearing for less than 100 times in the training dataset. In Table \ref{pp_freq}, extended models always achieve better performance in perplexity for low-frequency queries than baselines. 
It demonstrates that extended models are capable of tackling sparse data effectively.
\begin{table}
  \caption{Performance on perplexity for different query frequencies. Extended models achieve the greatest improvements when query frequency is less than 10 times in the training dataset. The more sparse the training data is, the greater improvements the extended models have.}
  \label{pp_freq}
  \begin{tabular}{c|cc|cc}
    \hline
    Query Freq.&UBM&vUBM-2&PBM&vPBM-2\\
    \hline
    1-10&1.389&\textbf{1.238 (+38.82\%)}&1.381&\textbf{1.228 (+40.16\%)}\\
    10-30&1.381&1.298 (+21.78\%)&1.375&1.284 (+24.27\%)\\
    30-100&1.374&1.330 (+11.76\%)&1.370&1.319 (+13.78\%)\\
    \hline
\end{tabular}
\end{table}
\subsubsection{Ranking Effectiveness}
In addition to data fitness, this paper also discusses if extended models handle sparsity better when it is evaluated by ranking effectiveness. Following \cite{mrr}, to compare different click models on their effectiveness in rankings, we use the relevance parameters in the click models and compare the accuracy in relevance prediction. Mean Reciprocal Rank (MRR) is such a metric to evaluate the performance of click models on their ranking effectiveness. 
Formally, given the test dataset with \textit{n} queries, the MRR is defined as:
\begin{equation}
\begin{aligned}
MRR = \frac{1}{n} \sum_{i=1}^{n} \frac{1}{rank_{i}}
\end{aligned}
\end{equation}
\par
The experiment results are given in the Table \ref{mrr}. From this table there comes the conclusion that when PBM and UBM incorporates \textit{vision bias} as a part of model hypothesis, it predicts the ranking better on low-frequency queries than their baseline models. It demonstrates that extended models can handle sparse data well.
\begin{table}
  \caption{Performance on MRR for different query frequencies. Extended models achieve further improvements when query frequency is less than 100 times in the training dataset. }
  \label{mrr}
  \begin{tabular}{c|cc|cc}
    \hline
    Query Freq.&UBM&vUBM-2&PBM&vPBM-2\\
    \hline
    1-10&0.817&0.835 (+2.20\%)&0.817&0.833 (+1.96\%)\\
    10-30&0.733&0.758 (+3.41\%)&0.733&0.759 (+3.55\%)\\
    30-100&0.665&0.677 (+1.80\%)&0.665&0.683 (+2.71\%)\\
  \hline
\end{tabular}
\end{table}

\subsection{Visualization Analysis}
\begin{figure*}
\centering  
\subfigure[Images  with a large or small vision  bias  corresponding to  the  same  query  "Yes,  beauty". The  left  one  has  colorful  texts  and brighter color.]{
 \label{fig:e}     
\includegraphics[height=0.35\columnwidth]{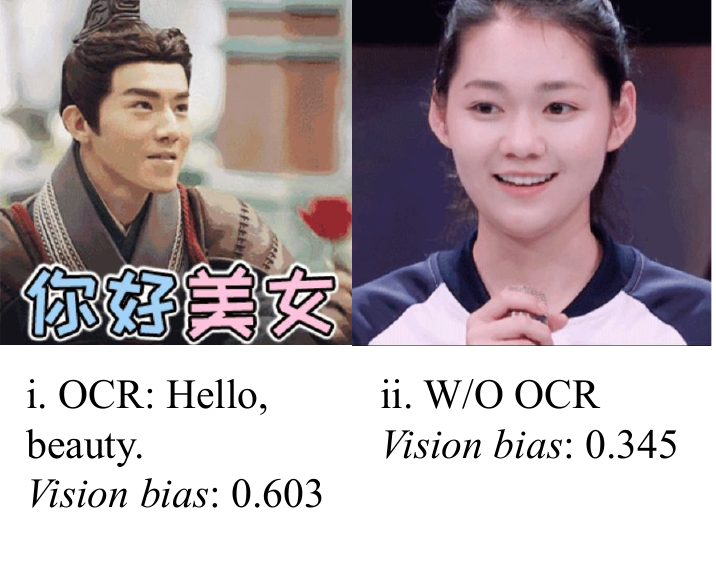}  
}
\quad
\subfigure[Images with a large or small vision bias corresponding to the same query "Try hard". The left one has a background of contrasting colors.]{ 
\label{fig:f}     
\includegraphics[height=0.35\columnwidth]{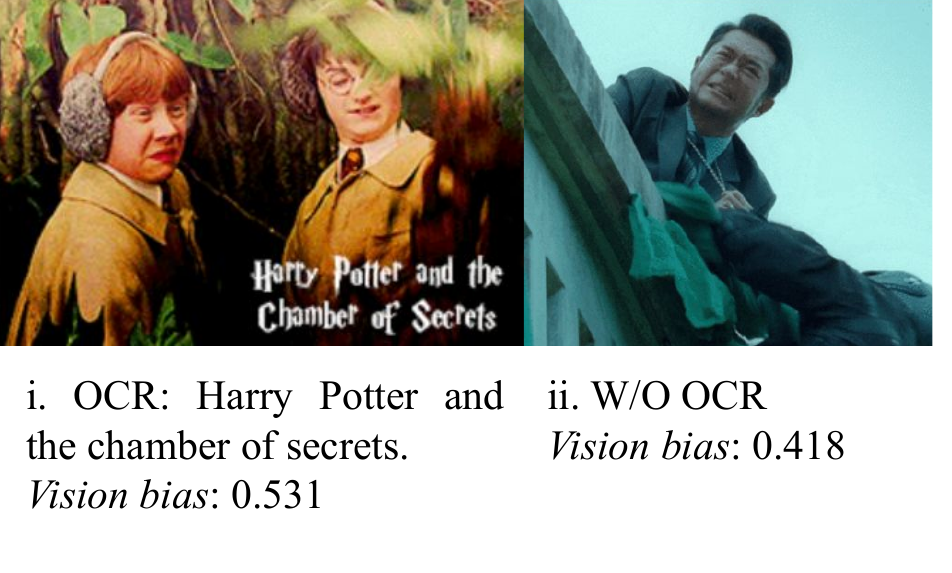}  
}   
\quad
\subfigure[Images with a large or small vision bias corresponding to the same query "Thanks for your love". The left one has concise but expressive shape.]{ 
\label{fig:g}     
\includegraphics[height=0.35\columnwidth]{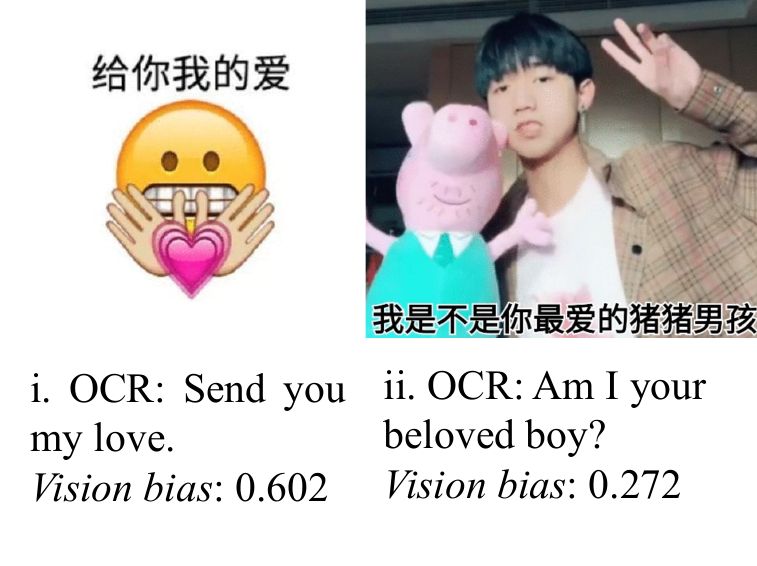} 
}  
\caption{Images with a large or small vision bias corresponding to the same query.} 
\label{qd_bias}   
\end{figure*}   

This section studies how \textit{vision bias} is distributed. 
In order to find if \textit{vision bias} has physical meaning in real-world image-oriented search engines, we first compare pairs of images corresponding to the same query with a large and a small \textit{vision bias}. We then analyze what characteristics in images influence the value of \textit{vision bias} by discussing images with higher and lower \textit{vision biases} that do not correspond to the same query.
\par
The images in the first pair of Figure \ref{qd_bias} differ in two ways. First, the one with larger bias has brighter color. Second, the one with larger bias has texts written in colors contrasting with its background. Texts and brightness can be factors that influence image vision bias. In Figure \ref{qd_bias}(b), the main difference between images mainly lays on colors. The image with smaller bias, compared with the other one, has a background of similar colors, which declines its attractiveness to users. The last pair in Figure \ref{qd_bias} shows two totally different images. The emoji one has concise shapes, but it is easy to understand and its colors are bright. The image with smaller bias looks dim, and the picture quality is not quite clear. These factors can be the reason for the difference between their biases.
  
\begin{figure}
\centering  
\includegraphics[height=0.42\columnwidth]{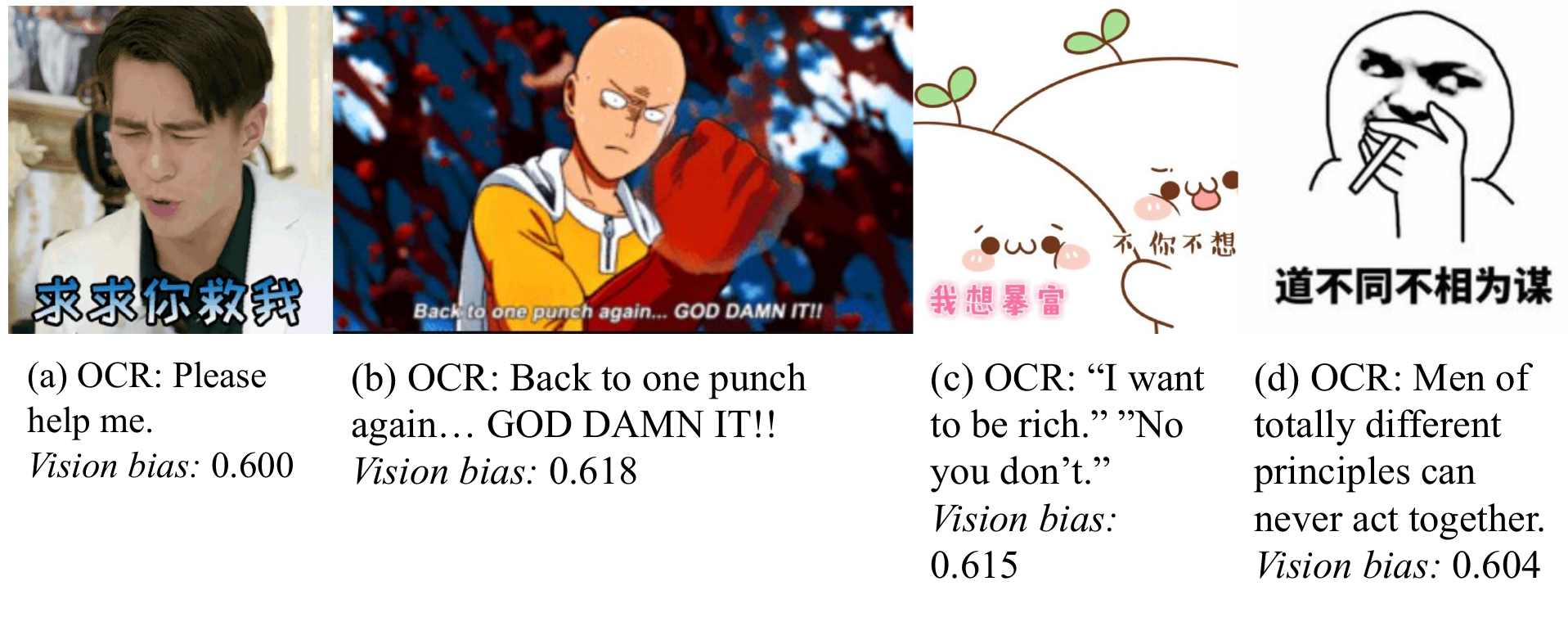}
 
\caption{Images with large vision biases. They have bright or contrasting background color, concise shapes and colorful texts.}
\label{large} 
\end{figure} 

\begin{figure}
\centering  
\includegraphics[height=0.34\columnwidth]{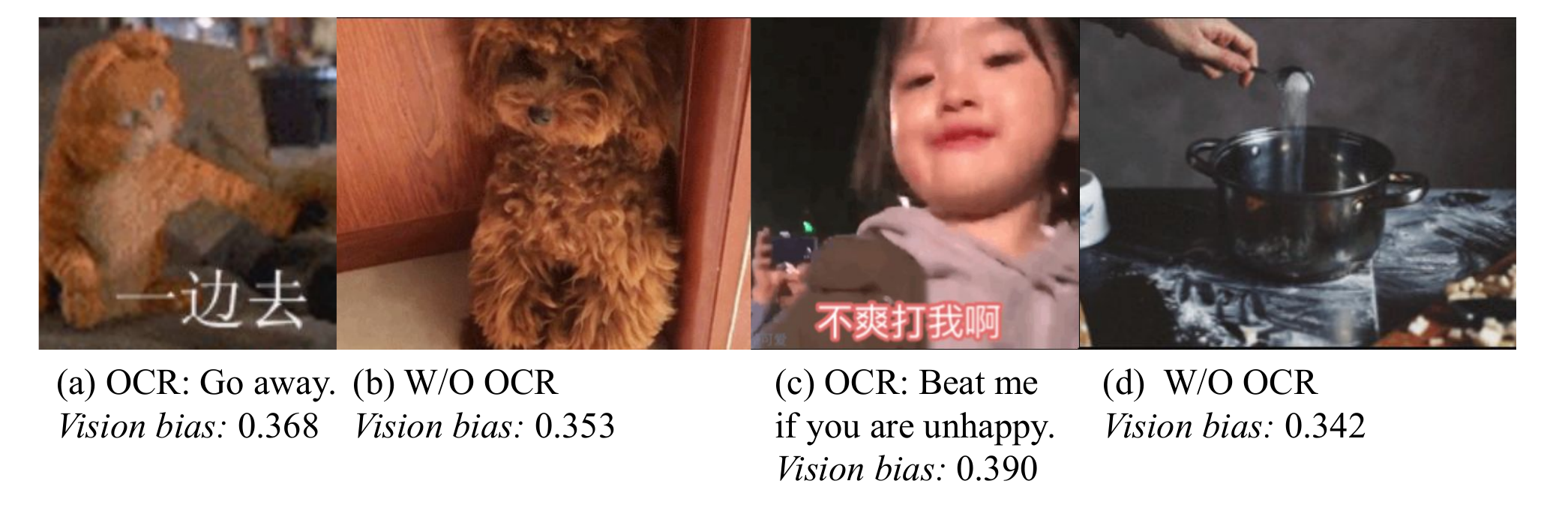}

\caption{Images with small vision biases. They look dim and not eye-catching.}
\label{small}  
\end{figure} 

\par
The images in Figure \ref{large} and Figure \ref{small} differ a lot. We find that the former with contrasting colors has larger \textit{vision bias}. The backgrounds in Figure \ref{large}(c) and Figure \ref{large}(d) are clean, thus making the cartoon characters stand out. Although Figure \ref{large}(b), Figure \ref{small}(b) and Figure \ref{small}(d) have dark background, colors of character in Figure \ref{large}(b) are more bright and eye-catching, which can be another way for an image to get higher \textit{vision bias}. As for the texts in the images, comparisons can be made between Figure \ref{large}(a), Figure \ref{large}(d) and Figure \ref{small}(a), Figure \ref{small}(c). In this comparison, images with larger \textit{vision bias} have texts very different from background. Compared with texts in similar color to  backgrounds, this can be another possible reason that leads to larger \textit{vision bias}. Besides, Figure \ref{large}(c) and Figure \ref{large}(d) are different from others, because they are cartoon characters with quite simple lines and shapes. Among other images, such
concise characters make them stand out.
\par
In fact, \textit{vision bias} should not be attributed to enumerable factors. Our model can understand and catch  it more deeply and comprehensively from high-dimensional visual features.

%% file: appendix.tex
\clearpage
\section*{Supplemental Material}
\appendix
\section{EM algorithm}

By giving the specific formula, we explain the E-step and the M-step of the standard EM algorithm to estimate the parameter $\sigma_{d}$. 

\subsection{E-step}
\begin{displaymath}
\begin{aligned}
P \left( E_{r}=0 , C_{<r_{d}}\mid \textbf{C} \right) &=P \left( E_{r}=0 \mid C_{<r_{d}}, \textbf{C} \right) \cdot P \left(C_{<r_{d}} \mid \textbf{C}\right)\\
&=P \left( E_{r}=0 \mid \textbf{C} \right)\\
&= c_{d} \cdot P \left( E_{r}=0 \mid C_{r}=1 \right) \\
&+ \left(1 - c_{d}\right) \cdot P \left( E_{r}=0 \mid C_{r}=0 \right) 
\end{aligned}
\label{indep}
\end{displaymath}
\begin{displaymath}
\begin{aligned}
P \left( E_{r}=0 \mid C_{r}=1 \right) &= 
\frac{P \left(E_{r}=0, C_{r}=1 \right)}{P \left(C_{r}=1 \right)}\\
&=\frac{P \left( E_{r}=0, \widetilde{E}=1, R_{r}=1, C_{r}=1 \right)}{P \left( C_{r}=1 \right)} \\
&=\frac{\alpha_{_{qd}}\sigma_{d}\left(1-\gamma_{_{rr'}}\right)}{\alpha_{_{qd}}\left(\gamma_{_{rr'}}+\sigma_{d}\left(1-\gamma_{_{rr'}}\right)\right)} \\ 
\end{aligned}
\end{displaymath}
\begin{displaymath}
\begin{aligned}
&\quad P \left( E_{r}=0 \mid C_{r}=0 \right) \\
&=
\frac{P \left(E_{r}=0, C_{r}=0 \right)}{P \left(C_{r}=0 \right)}\\
&=\frac{P \left(E_{r}=0, R_{r}=0, C_{r}=0 \right) + P \left(E_{r}=0, R_{r}=1, C_{r}=0 \right)}{P \left(C_{r}=0 \right)}\\
&=\frac{\left(1-\gamma_{_{rr'}}\right)\left(1-\alpha_{_{qd}}\right) + \alpha_{_{qd}}\left(1-\gamma_{_{rr'}}\right)\left(1-\sigma_{d}\right)}{1-\alpha_{_{qd}}\left(\gamma_{_{rr'}} + \sigma_{d}\left(1-\gamma_{_{rr'}}\right)\right)}
\end{aligned}
\end{displaymath}
Equation \ref{indep} holds because the click events at different ranks are independent or conditionally independent. Similarly, the prior distribution of the event $\widetilde{E}=1$ and $E=0$ can be derived as follows.
\begin{displaymath}
\begin{aligned}
P \left( \widetilde{E}_{r}=1, E_{r}=0, \textbf{$C_{<r_{d}}$} \mid \textbf{C}\right) &= P \left( \widetilde{E}_{r}=1, E_{r}=0 \mid \textbf{$C_{<r_{d}}$}, \textbf{C}\right)\\
&\cdot P \left(\textbf{$C_{<r_{d}}$} \mid \textbf{C}\right) \\
&=P \left(  \widetilde{E}_{r}=1, E_{r}=0 \mid \textbf{C}\right) \\
&=c_{d} \cdot P \left(  \widetilde{E}_{r}=1, E_{r}=0 \mid C_{r}=1 \right) \\
&+ \left(1 - c_{d}\right) \cdot P \left(  \widetilde{E}_{r}=1, E_{r}=0 \mid C_{r}=0 \right)
\end{aligned}
\end{displaymath}
\begin{displaymath}
\begin{aligned}
P \left(  \widetilde{E}_{r}=1, E_{r}=0 \mid C_{r}=1 \right) &= \frac{P \left(  \widetilde{E}_{r}=1, E_{r}=0, C_{r}=1 \right)}{P \left(C_{r}=1 \right)} \\
&= \frac{\alpha_{_{qd}}\sigma_{d}\left(1-\gamma_{_{rr'}}\right)}{\alpha_{_{qd}}\left(\gamma_{_{rr'}} + \sigma_{d}\left(1-\gamma_{_{rr'}}\right)\right)}
\end{aligned}
\end{displaymath}
\begin{displaymath}
\begin{aligned}
P \left(  \widetilde{E}_{r}=1, E_{r}=0 \mid C_{r}=0 \right) &= \frac{P \left(  \widetilde{E}_{r}=1, E_{r}=0, C_{r}=0 \right)}{P \left(C_{r}=0 \right)} \\
&=\frac{\sigma_{d}\left(1-\gamma_{_{rr'}}\right)\left(1-\alpha_{_{qd}}\right)}{1-\alpha_{_{qd}}\left(\gamma_{_{rr'}} + \sigma_{d}\left(1-\gamma_{_{rr'}}\right)\right)}
\end{aligned}
\end{displaymath}
Above are the formula for the parameter $\sigma_{d}$ to optimize the model in the next step.
\subsection{M-step}
\par
In the Maximization step, the standard EM algorithm updates the parameters in iteration $t+1$ by maximizing the likelihood of the posterior probabilities obtained in the Expectation step given the click data. For example, to update the parameter $\sigma_{d}$, we use the following equation
\begin{displaymath}
\begin{aligned}
\sigma_{d}^{t+1}&=\frac{\sum_{s \in S_{q}}P \left( \widetilde{E}_{r}=1, E_{r}=0, C_{<r_{d}} \mid \textbf{C}\right)}{\sum_{s \in S_{q}}P \left( E_{r}=0 , C_{<r_{d}}\mid \textbf{C} \right)}
\end{aligned}
\label{mstep}
\end{displaymath}
$S_{q}$ represents the search session where $q$ is given as the query. According to the formulas above, finally the parameter $\sigma_{d}$ in the iteration $t+1$ can be updated based on parameters obtained in the iteration $t$.
\par
To summarize, all probabilities of hidden variables in the E-step in the extended UBM model can be computed by the following formula.
\begin{displaymath}
\begin{aligned}
P\left(R_{r}=1 \mid \textbf{C} \right) = c_{d}+\left(1-c_{d}\right)\alpha_{qd}^{t}\frac{1-\left(\gamma_{rr'}+\sigma_{d}-\gamma_{rr'}\sigma_{d}\right)}{1-\alpha_{qd}\left(\gamma_{rr'}+\left(1-\gamma_{rr'}\right)\sigma_{d}\right)}
\end{aligned}
\end{displaymath}
\begin{displaymath}
\begin{aligned}
P\left(E_{r}=1, C_{<r_{d}} \mid \textbf{C} \right) &=c_{d}\frac{\gamma_{rr'}}{\gamma_{rr'}+\sigma_{d}\left(1-\gamma_{rr'}\right)}\\
&+\left(1-c_{d}\right)\frac{\gamma_{rr'}\left(1-\alpha_{qd}\right)}{1-\alpha_{qd}\left(\gamma_{rr'}+\left(1-\gamma_{rr'}\right)\sigma_{d}\right)}
\end{aligned}
\end{displaymath}
\begin{displaymath}
\begin{aligned}
P \left( \widetilde{E}_{r}=1, E_{r}=0, C_{<r_{d}} \mid \textbf{C}\right) &=c_{d}\frac{\sigma_{d}\left(1-\gamma_{rr'}\right)}{\gamma_{rr'}+\left(1-\gamma_{rr'}\right)\sigma_{d}}\\
&+\left(1-c_{d}\right)\frac{\sigma_{d}\left(1-\gamma_{rr'}\right)\left(1-\alpha_{qd}\right)}{1-\alpha_{qd}\left(\gamma_{rr'}+\left(1-\gamma_{rr'}\right)\sigma_{d}\right)}
\label{prob}
\end{aligned}
\end{displaymath}
\par
Accordingly, the M-step uses the following formulas to update all the parameters in the extended UBM model. We need to set up initial values for the parameters at iteration 0.
\begin{displaymath}
\begin{aligned}
\alpha_{_{qd}}^{t+1} = \frac{\sum_{s \in S_{q}} \left(c_{d}+\left(1-c_{d}\right)\alpha_{qd}^{t}\frac{1-\left(\gamma_{rr'}^{t}+\sigma_{d}^{t}-\gamma_{rr'}^{t}\sigma_{d}^{t}\right)}{1-\alpha_{qd}^{t}\left(\gamma_{rr'}^{t}+\left(1-\gamma_{rr'}^{t}\right)\sigma_{d}^{t}\right)} \right)}{\sum_{s \in S_{q}} 1}
\label{alpha2}
\end{aligned}
\end{displaymath}
\begin{displaymath}
\begin{aligned}
\gamma_{rr'}^{t+1}=\frac{\sum_{s \in S_{q}}\left(c_{d}\frac{\gamma_{rr'}^{t}}{\gamma_{rr'}^{t}+\sigma_{d}^{t}\left(1-\gamma_{rr'}^{t}\right)}+\left(1-c_{d}\right)\frac{\gamma_{rr'}^{t}\left(1-\alpha_{qd}^{t}\right)}{1-\alpha_{qd}^{t}\left(\gamma_{rr'}^{t}+\left(1-\gamma_{rr'}^{t}\right)\sigma_{d}^{t}\right)}\right)}{\sum_{s \in S_{q}} 1}
\label{gamma2}
\end{aligned}
\end{displaymath}
\begin{displaymath}
\begin{aligned}
\sigma_{d}^{t+1}=\frac{\sum_{s \in S_{q}}\left(c_{d}\frac{\sigma_{d}^{t}\left(1-\gamma_{rr'}^{t}\right)}{\gamma_{rr'}^{t}+\left(1-\gamma_{rr'}^{t}\right)\sigma_{d}^{t}}+\left(1-c_{d}\right)\frac{\sigma_{d}^{t}\left(1-\gamma_{rr'}^{t}\right)\left(1-\alpha_{qd}^{t}\right)}{1-\alpha_{qd}^{t}\left(\gamma_{rr'}^{t}+\left(1-\gamma_{rr'}^{t}\right)\sigma_{d}^{t}\right)}\right)}{\sum_{s \in S_{q}}\left(c_{d}\frac{\sigma_{d}^{t}\left(1-\gamma_{rr'}^{t}\right)}{\gamma_{rr'}^{t} + \left(1-\gamma_{rr'}^{t}\right)\sigma_{d}^{t}}+\left(1-c_{d}\right)\frac{\left(1-\gamma_{rr'}^{t}\right)\left(1-\sigma_{d}^{t}\alpha_{qd}^{t}\right)}{1-\alpha_{qd}^{t}\left(\gamma_{rr'}^{t}+\left(1-\gamma_{rr'}^{t}\right)\sigma_{d}^{t}\right)}\right)}
\end{aligned}
\end{displaymath}  